\documentclass[journal]{IEEEtran}
\usepackage{silence}
\usepackage{amsmath,amssymb}
\usepackage{graphicx}
\usepackage[table]{xcolor}
\usepackage{booktabs}
\usepackage{array}
\usepackage{titlesec}
\usepackage[T1]{fontenc}
\usepackage[utf8]{inputenc}
\usepackage{url}
\usepackage[font=small,labelfont=bf]{caption}
\usepackage{float}
\definecolor{headerblue}{RGB}{220,230,245}
\definecolor{rowgray}{RGB}{248,249,251}

\renewcommand{\thesection}{\arabic{section}}
\renewcommand{\thesubsection}{\thesection.\arabic{subsection}}

\titleformat{\section}
  {\normalfont\normalsize\bfseries}
  {\thesection.}{0.6em}{}

\titleformat{\subsection}
  {\normalfont\small\bfseries}
  {\thesubsection.}{0.5em}{}

\begin{document}

\title{A Proof-of-Concept Simulation-Driven Digital Twin Framework for Decision-Aware Diabetes Modeling}
\author{
\Large Zarrin Monirzadeh\\[0.6em]
\normalsize Software \& Data Engineer | ML \& AI Systems\\
\normalsize Portland, OR, USA
}
\maketitle

\begin{abstract}
This paper presents a proof-of-concept digital twin framework for simulation-driven diabetes modeling using benchmark clinical data, synthetic temporal augmentation, and illustrative continuous glucose monitoring (CGM) analysis.

Unlike traditional predictive models, the framework focuses on generating interpretable simulated trajectories rather than clinically validated outcomes. Evaluation is conducted using a public dataset combined with controlled synthetic scenarios to illustrate temporal behavior and intervention effects.

Results illustrate the feasibility of integrating prediction with counterfactual simulation for decision-aware analysis. This work does not claim clinical readiness but provides a foundation for future research on simulation-driven digital twin systems in healthcare.
\end{abstract}


\section{Introduction}
Diabetes mellitus is a major global health challenge with significant impact on morbidity, mortality, and healthcare systems. It is a heterogeneous group of metabolic disorders rather than a single disease, with substantial variation in underlying physiology and treatment requirements. Type 1 diabetes involves autoimmune destruction of pancreatic beta cells and requires lifelong insulin therapy, while Type 2 diabetes is associated with insulin resistance and progressive metabolic dysfunction. Gestational diabetes occurs during pregnancy and introduces additional maternal and fetal risks.

Artificial intelligence has shown strong performance in diabetes risk prediction, stratification, and glucose forecasting~\cite{ref1,ref2,ref3,ref9,ref23,ref24,ref25}. Advances in machine learning and continuous glucose monitoring have enabled accurate estimation of short-term glycemic trends. However, most existing systems remain prediction-oriented. They estimate future outcomes but provide limited guidance on what actions should be taken or how interventions may alter those outcomes.

This limitation reflects a broader gap in current artificial intelligence systems in healthcare. While predictive models provide estimates of future outcomes, they do not directly support decision-making under intervention. In contrast, clinically meaningful intelligence requires the ability to evaluate alternative actions and their potential effects. This distinction motivates a transition from predictive artificial intelligence toward decision-aware systems, in which models are designed not only to forecast outcomes but to support actionable, intervention-driven reasoning.

In practice, diabetes management is inherently decision-driven. Patients and clinicians continuously evaluate interventions such as diet, physical activity, and treatment adjustments under uncertainty. Therefore, intelligent systems must move beyond passive prediction and support actionable decision-making.

Digital twin technology provides a promising computational foundation for transitioning from static prediction systems toward adaptive, patient-specific, and intervention-aware clinical decision frameworks~\cite{ref4,ref5,ref6,ref7,ref8,ref20,ref21,ref22,ref26}. A digital twin is a dynamic computational representation of a physical system capable of estimating internal state, predicting future behavior, and evaluating interventions. In healthcare, this enables patient-specific modeling and simulation-based decision support, which is particularly relevant for diabetes due to its continuous and measurable nature.

Recent research in healthcare digital twins has increasingly emphasized patient-specific simulation, temporal modeling, and intervention-aware analysis~\cite{ref4,ref5,ref6,ref7,ref8,ref20,ref21,ref22,ref26}. These studies highlight the growing importance of integrating predictive modeling with individualized decision support in chronic disease management.

Despite this potential, current research remains fragmented and lacks a unified framework that integrates temporal learning, causal reasoning, and multi-type disease modeling. Many studies are either conceptual or limited to narrow experimental settings without reproducible evaluation.

However, many of these approaches focus either on prediction or conceptual modeling without providing a simplified, reproducible architecture for simulation-based decision analysis. This work positions itself as a proof-of-concept framework emphasizing reproducibility and integration of temporal modeling and counterfactual simulation.

This paper addresses these limitations by introducing a personalized digital twin framework for multiple diabetes types. The proposed approach integrates heterogeneous data, temporal modeling, and counterfactual simulation to support decision-oriented analysis and interpretable comparison of intervention strategies.

The primary contribution of this work is the integration of temporal modeling, causal reasoning, and counterfactual simulation within a unified decision-oriented framework.

The main contributions of this paper are as follows:
\begin{itemize}
\item A modular digital twin architecture adaptable across Type 1, Type 2, and gestational diabetes.
\item A transition from correlation-based prediction to decision-aware modeling, enabling explicit evaluation of intervention strategies through causal and counterfactual reasoning.
\item A reproducible proof-of-concept evaluation using an open dataset and simulated scenarios.
\item A discussion of practical deployment considerations, including interpretability, safety, privacy, and validation.
\end{itemize}

\section{Clinical Computing Foundations}
\subsection{Diabetes as a Multi-Type Control Problem}
From a computational perspective, diabetes management can be formulated as a dynamic control problem under uncertainty. The underlying physiological state evolves over time, observations are incomplete and noisy, external inputs occur frequently, and interventions produce delayed and context-dependent effects.

In Type 1 diabetes, the primary challenge is short-term glucose regulation under the combined influence of insulin administration, food intake, physical activity, stress, and circadian variability. In Type 2 diabetes, longer-term disease progression becomes more prominent, including medication adherence, weight dynamics, and behavioral patterns. Gestational diabetes introduces pregnancy-specific hormonal changes and reduced tolerance for uncertainty due to maternal and fetal safety requirements.

These findings highlight the limitations of static prediction models that cannot continuously adapt to evolving physiological states or dynamically evaluate intervention effects. A clinically useful system must capture the evolving disease state, infer latent physiological structure, and estimate the effects of candidate interventions. Prior work has demonstrated that time-series models, including recurrent, transformer-based, and hybrid deep learning approaches, can improve short-term glucose prediction~\cite{ref9,ref10,ref23,ref24,ref25,ref30,ref31,ref32}. However, predictive accuracy alone is not sufficient to meet clinical needs. The key question is not only what will happen next, but which action is most likely to lead to a safer and more desirable outcome.

\subsection{Digital Twins in Healthcare}
Medical digital twins are commonly defined as patient-specific computational models that integrate physiological knowledge, individualized data, and continuous state updating~\cite{ref4,ref5,ref6,ref7,ref8,ref20,ref21,ref22,ref26}. These systems enable simulation-based reasoning, treatment planning, and evaluation of interventions prior to real-world execution.

The literature highlights several key advantages, including improved personalization, the ability to conduct in silico experimentation, and the potential to reduce the cost and duration of clinical studies. At the same time, significant challenges remain, including limitations in data quality, interoperability, calibration, model drift, accountability, privacy, and ethical governance~\cite{ref7,ref8,ref13,ref21,ref22,ref26}.

Diabetes is particularly well suited to digital twin approaches due to the availability of continuous sensing and the presence of frequent, measurable interventions. Recent studies on digital twins for Type 1 diabetes emphasize the need for adaptive patient models capable of supporting insulin therapy, glucose prediction, and decision support~\cite{ref4}.

However, healthcare digital twins differ fundamentally from industrial systems. The system being modeled is biologically adaptive, partially observable, and strongly influenced by behavioral and contextual factors. The true patient state cannot be directly measured and must instead be inferred. Consequently, digital twins in healthcare must operate under uncertainty, rely on latent state estimation, and incorporate elements of causal structure. These characteristics increase the importance of interpretability, safety, and robustness.

\subsection{Why Causality Matters}
Many current artificial intelligence systems in healthcare learn predictive relationships without explicitly distinguishing causal structure from correlation. In the context of diabetes management, this limitation is particularly important. A model may identify that elevated glucose levels are associated with reduced physical activity, but such associations alone do not provide reliable guidance for intervention.

Effective decision support requires estimation of intervention effects. For example, it is necessary to answer questions such as how glucose levels are expected to change if physical activity is increased or if carbohydrate intake is reduced. Addressing these questions requires counterfactual reasoning rather than purely predictive modeling~\cite{ref13,ref18,ref19}.

A digital twin should therefore support explicit intervention analysis. Rather than reporting expected future values alone, it should enable statements such as: given the current patient state, reducing carbohydrate intake from 60 grams to 30 grams is expected to lower peak glucose within a defined time window. This capability distinguishes decision-oriented systems from purely predictive models and represents a central component of the framework proposed in this work.

\section{Related Work}
Research on artificial intelligence in diabetes has evolved along four primary directions. The first focuses on diagnosis, complication prediction, and clinical risk stratification using machine learning applied to structured clinical data~\cite{ref1,ref2,ref3,ref16,ref17,ref27}. The second centers on continuous glucose monitoring and time-series forecasting, where classical machine learning methods, recurrent neural networks, transformer-based models, and hybrid deep architectures have been used to predict short-term glucose dynamics~\cite{ref9,ref10,ref23,ref24,ref25,ref30,ref31,ref32}. The third direction emphasizes personalization, interpretability, and privacy-aware learning, reflecting the need for patient-specific models that can operate in realistic healthcare environments~\cite{ref11,ref12,ref14,ref35,ref36,ref37}. The fourth introduces digital twins as a broader paradigm for patient-specific simulation, state estimation, and individualized intervention analysis~\cite{ref4,ref5,ref6,ref7,ref8,ref20,ref21,ref22,ref26}.

In addition, prior work by the author on machine learning-based detection of violations in credit card systems highlights the broader relevance of anomaly detection, risk scoring, and decision support under uncertainty~\cite{ref15}. Although the application domain differs, the underlying methodological theme is similar: predictive models become more valuable when they support actionable decisions rather than static classification.

Early work by Contreras and Vehi~\cite{ref1} established the importance of decision support and automated therapy concepts in artificial intelligence systems for diabetes management. Kavakiotis et al.~\cite{ref2} further demonstrated the breadth of machine learning applications in diabetes, spanning diagnosis, prognosis, and treatment support across multiple data modalities. More recent studies have leveraged continuous glucose monitoring to enable richer temporal modeling. van Doorn et al.~\cite{ref9} showed that incorporating meal and physical activity information improves glucose forecasting performance. Bian et al.~\cite{ref10} reported that hybrid transformer and recurrent architectures can better capture temporal dependencies in Type 1 diabetes data, while Darpit et al.~\cite{ref11} extended this line of work toward federated personalization under privacy constraints. Collectively, these studies demonstrate that modern artificial intelligence methods can learn complex temporal and patient-specific patterns in diabetes-related data.

\begin{figure}[htbp]
\centering
\includegraphics[width=0.95\linewidth]{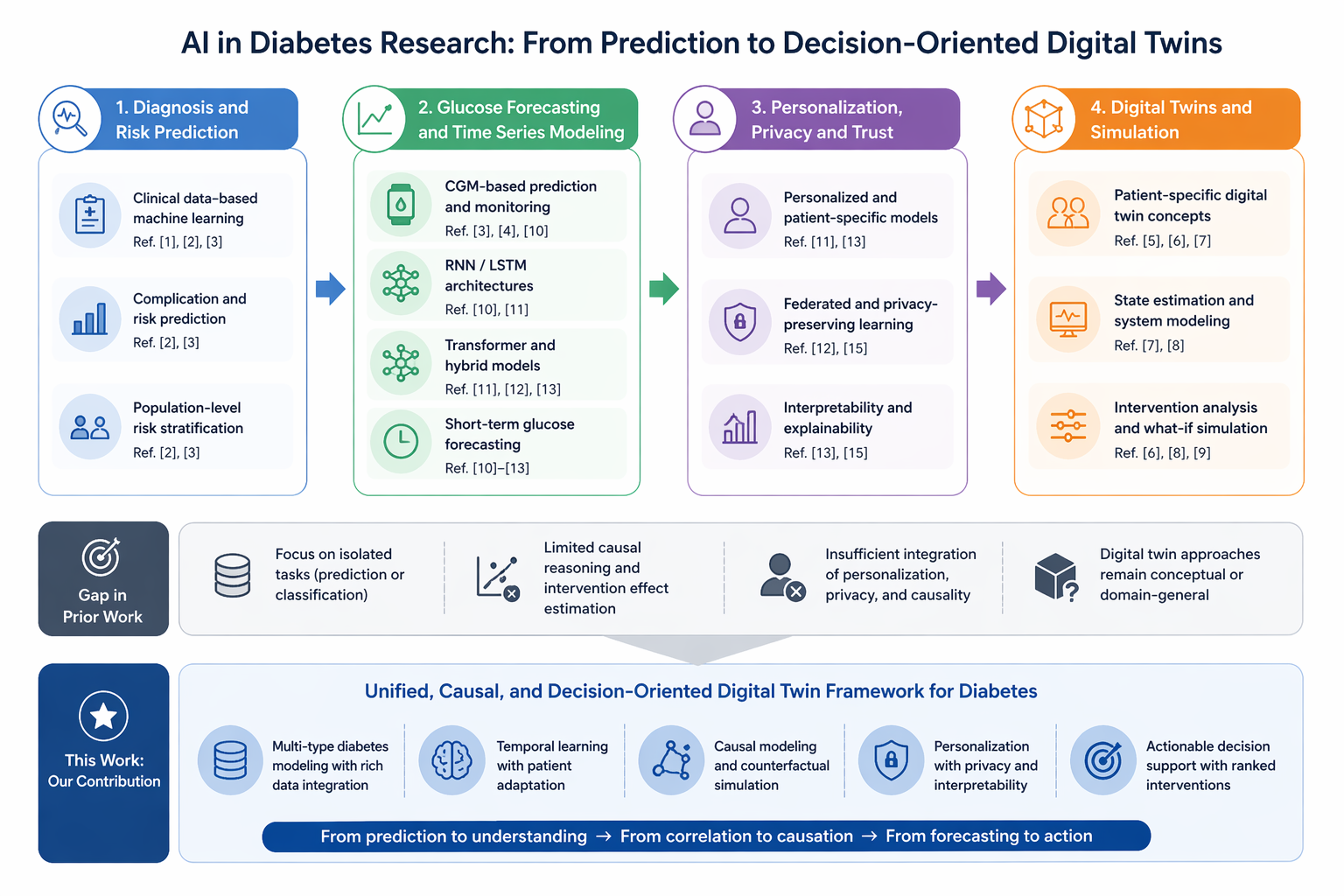}
\caption{Overview of artificial intelligence approaches in diabetes, highlighting the transition from prediction-focused models toward decision-oriented and simulation-based frameworks.}
\label{fig:related_work}
\end{figure}

Despite this progress, an important limitation remains. Most existing studies focus on prediction, classification, or risk scoring. While these approaches estimate future outcomes, they typically do not evaluate alternative actions or quantify the expected effects of interventions. In parallel, digital twin literature highlights simulation, personalization, and model-based reasoning~\cite{ref4,ref5,ref6,ref7,ref8}, but many contributions remain conceptual, domain-general, or insufficiently operationalized for diabetes-specific decision support.

The present work addresses this gap by integrating these research directions into a unified decision-oriented framework. Rather than introducing a fundamentally new learning algorithm, the contribution lies in a structured synthesis of components that are rarely combined within a single system. The proposed approach unifies multiple diabetes types modeling, temporal learning, causal reasoning, counterfactual intervention analysis, and reproducible proof-of-concept evaluation. In this way, the work moves beyond isolated prediction tasks toward a coherent framework for actionable decision support and future clinical translation.

\section{Causal Digital Twin Framework}
\subsection{Design Principles}
The proposed framework is guided by five core principles: patient specificity, multimodality, temporal continuity, counterfactual actionability, and interpretability under safety constraints. Patient specificity emphasizes adaptation to individual physiological characteristics rather than reliance on population-level averages. Multimodality reflects the integration of clinical, physiological, and behavioral signals. Temporal continuity captures longitudinal dependencies instead of isolated observations. Counterfactual actionability enables evaluation of alternative interventions, while interpretability under safety constraints ensures that recommendations remain explainable and clinically appropriate.

These principles align with established requirements for trustworthy healthcare artificial intelligence, including transparency, interpretability, safety, and clinical relevance~\cite{ref16,ref17,ref21,ref35,ref36,ref37}.

\subsection{System Overview}
The proposed framework follows a modular pipeline consisting of data ingestion, state representation, predictive modeling, and counterfactual simulation. Clinical, physiological, and behavioral inputs are first harmonized and transformed into structured representations. A latent state representation is then constructed to capture patient-specific dynamics, which are subsequently used for prediction and simulation of intervention outcomes.

\begin{figure}[htbp]
\centering
\includegraphics[width=0.95\linewidth]{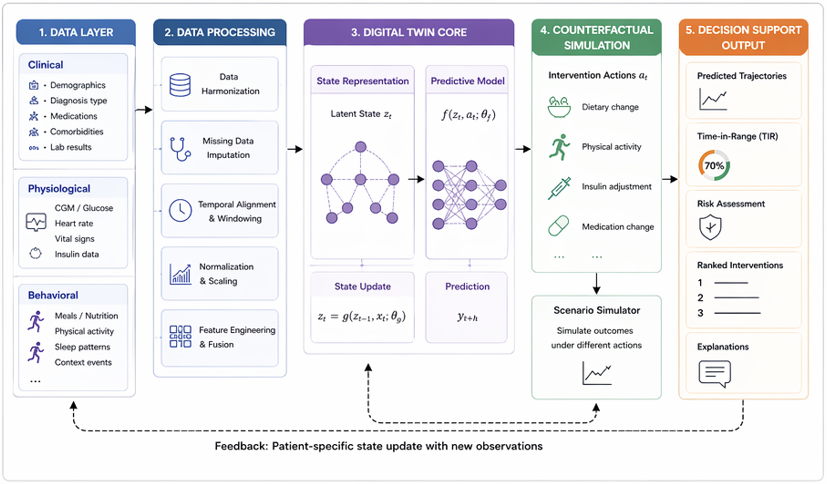}
\caption{Proposed digital twin architecture illustrating the flow from data ingestion and preprocessing to latent state representation, predictive modeling, and counterfactual simulation for intervention analysis.}
\label{fig:architecture}
\end{figure}

\subsection{Data Layer}
The input layer integrates heterogeneous variables across three categories: clinical, physiological, and behavioral. Clinical variables include demographic information, diagnosis type, medication regimen, and comorbidities. Physiological variables include glucose measurements and, when available, additional biometric signals. Behavioral variables include meal intake, physical activity, sleep patterns, and other contextual factors.

In the current implementation, multimodal integration is demonstrated conceptually, while experimental validation is conducted on structured benchmark data.

Due to differences in sampling rates, noise levels, and data reliability, preprocessing involves temporal alignment, normalization, missing-value imputation, and feature fusion. For real-time applications, a rolling time window is employed to capture recent dynamics and maintain temporal consistency.

\subsection{State Representation}
Let $x_t$ denote the observed input vector at time $t$, and let $z_t$ denote the latent patient state. The system evolves according to
\[
z_t = g(z_{t-1}, x_t; \theta_g),
\]
and generates predictions as
\[
y_{t+h} = f(z_t, a_t; \theta_f),
\]
where $a_t$ represents a candidate intervention and $h$ denotes the prediction horizon. The latent state $z_t$ encodes patient-specific physiological context accumulated over time and enables simulation under alternative intervention scenarios.

This formulation is consistent with state-space modeling and sequential learning perspectives commonly used in temporal modeling and control-oriented machine learning~\cite{ref30,ref31,ref32,ref33}.

\subsection{Learning Models}
The predictive mapping $f(\cdot)$ is implemented using supervised learning models. For gradient boosting, the prediction is given by
\[
\hat{y} = \sum_{m=1}^{M} \gamma_m h_m(x),
\]
where $h_m(x)$ denotes the $m$-th weak learner and $\gamma_m$ represents its corresponding weight.

\begin{figure}[htbp]
\centering
\includegraphics[width=0.9\linewidth]{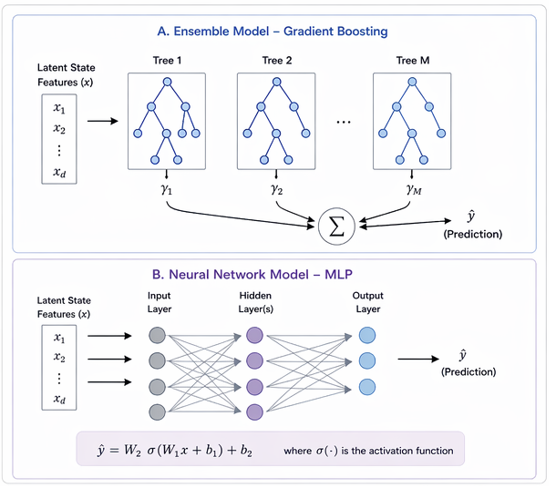}
\caption{Mapping between input features and predictive models, including both classical machine learning methods and neural network-based approaches within the digital twin framework.}
\label{fig:model_mapping}
\end{figure}

Gradient boosting and tree-based ensemble methods are widely used for structured data, as they effectively capture nonlinear feature interactions while maintaining strong performance on moderate-sized datasets~\cite{ref28,ref29}.

For neural network models, the prediction can be expressed as
\[
\hat{y} = W_2 \sigma(W_1 x + b_1) + b_2,
\]
where $W_1$ and $W_2$ are weight matrices, $b_1$ and $b_2$ are bias terms, and $\sigma(\cdot)$ is a nonlinear activation function.

These formulations provide concrete and reproducible instantiations of the predictive component within the digital twin framework. Neural network-based extensions can further incorporate recurrent and attention-based architectures to model temporal dependencies in sequential data~\cite{ref30,ref31,ref32}.

\subsection{Temporal Core and Causal Structure}
The framework supports temporal modeling through architectures such as long short-term memory (LSTM) networks and transformer-based models. A causal graph is used to encode relationships among key variables, including nutrition, physical activity, insulin administration, and glucose dynamics. This structure ensures that intervention simulations follow plausible physiological pathways rather than arbitrary feature perturbations.

The temporal core captures sequential dependencies in patient data, while the causal structure provides a principled basis for intervention analysis. Together, these components enable consistent modeling of both temporal dynamics and intervention effects~\cite{ref18,ref19,ref30,ref31,ref32}.

\subsection{Counterfactual Simulation}
The simulation engine evaluates multiple intervention scenarios, including dietary modifications, physical activity, and treatment adjustments. For each scenario, the system predicts glucose trajectories and computes clinically relevant metrics such as peak glucose levels and time-in-range. Candidate interventions are then ranked using a utility function that balances effectiveness and safety considerations.

\begin{figure}[htbp]
\centering
\includegraphics[width=0.9\linewidth]{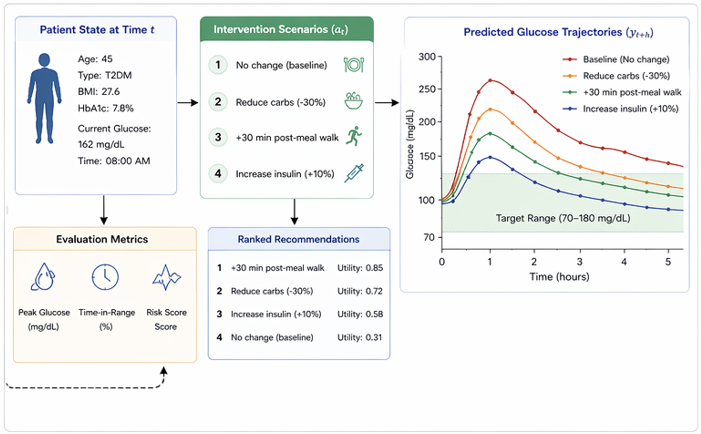}
\caption{Illustration of the decision-support workflow within the digital twin framework. The system evaluates multiple intervention scenarios and generates predicted glucose trajectories, enabling comparison of outcomes and ranking of candidate interventions.}
\label{fig:simulation}
\end{figure}

This figure illustrates the conceptual workflow of the decision-support process. Quantitative results derived from simulated trajectories are presented in Table~\ref{tab:counterfactual} and further visualized in Fig.~\ref{fig:counterfactual_plot}. The formulation is grounded in causal inference principles and aligns with established decision-support methodologies in artificial intelligence~\cite{ref18,ref19,ref33}.

\subsection{Implementation Details}
The proposed digital twin framework is implemented using a combination of classical machine learning methods and extensible deep learning components.

For the predictive layer, multiple supervised models are evaluated to ensure robustness across modeling assumptions, including gradient boosting regression, random forest regression, linear regression, and multilayer perceptron (MLP) models. Gradient boosting is selected as the primary baseline due to its strong empirical performance on structured medical datasets and its ability to capture nonlinear feature interactions.

All models are implemented in Python using the Scikit-learn library, with NumPy and Pandas used for data preprocessing, feature transformation, and evaluation. The implementation is designed to be fully reproducible and dataset-agnostic.

The framework supports integration with temporal deep learning architectures for modeling longitudinal patient dynamics. In particular, long short-term memory (LSTM) networks and transformer-based models can be incorporated to extend the latent state transition function defined in Section 4.3. These components can be implemented using standard deep learning libraries such as TensorFlow or PyTorch.

All regression and classification experiments were conducted using Scikit-learn implementations with fixed random seeds for reproducibility. For Gradient Boosting, 100 estimators with a learning rate of 0.1 were used. Random Forest models used 100 trees. The multilayer perceptron (MLP) model used a single hidden layer with ReLU activation and Adam optimization. Training and evaluation were performed using an 80:20 train-test split.

The counterfactual simulation engine operates by perturbing actionable input variables within clinically feasible ranges, including carbohydrate intake, physical activity duration, and intervention timing. For each modified scenario, the predictive model generates a corresponding trajectory, enabling comparison across alternative interventions. This hybrid design combines rule-based constraints with data-driven prediction to maintain both physiological plausibility and modeling flexibility.

\begin{figure}[htbp]
\centering
\includegraphics[width=0.95\linewidth]{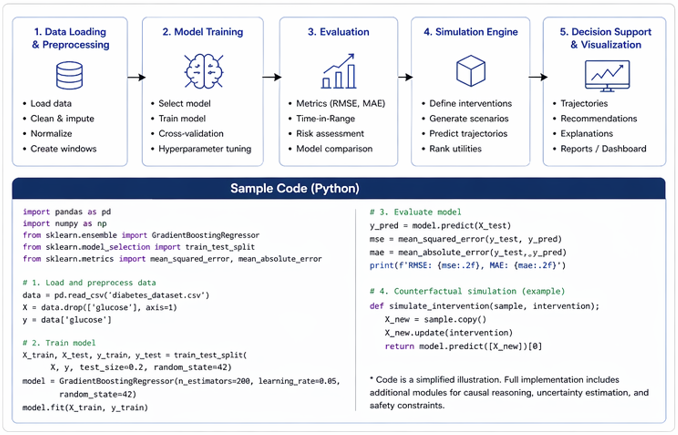}
\caption{Implementation workflow showing data preprocessing, model training, evaluation, and generation of simulated intervention scenarios for decision-support analysis.}
\label{fig:code}
\end{figure}

The selected modeling tools are consistent with established practices in machine learning and deep learning, including ensemble methods, neural networks, recurrent models, transformer architectures, and explainability techniques~\cite{ref28,ref29,ref30,ref31,ref32,ref35,ref36}.

The implementation prioritizes transparency and reproducibility. All experiments reported in this study are conducted using locally implemented models without reliance on external proprietary systems. Although the framework is compatible with modern large language model APIs for explanation or user interaction layers, such components are not required for the core predictive and simulation functionality and are not used in the reported experiments.

The full implementation, including reproducible code, generated datasets, experimental outputs, and visualization assets, is publicly available in a structured repository~\cite{ref41}.
\section{Materials and Experimental Methods}
\subsection{Proof-of-Concept Objective}
The purpose of the experimental evaluation is to assess the feasibility of the proposed digital twin framework under controlled and reproducible conditions. Rather than claiming clinical performance, the evaluation focuses on validating two core components of the system: the predictive modeling layer and the counterfactual simulation capability.

To this end, the study combines a benchmark regression experiment using a publicly available dataset with a simulated counterfactual analysis that illustrates decision-support behavior. The benchmark experiment evaluates whether the predictive models can learn meaningful relationships from diabetes-related data, while the simulation component demonstrates how the framework supports comparison of alternative interventions.

\subsection{Open Benchmark Dataset}
For reproducibility, the regression experiment uses the diabetes dataset provided in the Scikit-learn library. This dataset contains 442 samples with 10 standardized covariates and a continuous target variable representing disease progression.

The use of open benchmark data supports transparency and reproducibility, which are essential requirements in clinical artificial intelligence research~\cite{ref16,ref17,ref27}.

Although the dataset does not include time-series glucose measurements and therefore does not directly support real-time clinical evaluation, it provides a well-established benchmark for evaluating regression performance under transparent conditions.

The dataset is divided into training and test sets using an 80:20 split with a fixed random seed to ensure reproducibility.

\subsection{Dataset Construction and Temporal Augmentation}
To support the evaluation of the proposed framework, a hybrid data construction strategy was adopted. The baseline data is derived from the diabetes dataset provided by the Scikit-learn library, which offers a standardized and reproducible benchmark for regression tasks. This dataset contains structured clinical features and a continuous outcome variable representing disease progression.

While suitable for evaluating predictive performance, the dataset does not include longitudinal glucose measurements or explicit intervention variables. As a result, it cannot directly support temporal modeling or decision-oriented simulation.

To address this limitation, a simple temporal augmentation process was introduced. Rather than attempting to replicate full clinical realism, the objective was to construct controlled and interpretable time-dependent scenarios that approximate plausible glucose dynamics.

Each observation was extended into a short temporal sequence, and additional variables representing carbohydrate intake, physical activity, and intervention timing were incorporated within clinically reasonable ranges. Small stochastic variations were added to reflect natural variability and avoid deterministic behavior.

This augmentation enables the framework to simulate how different interventions may influence outcomes over time, which is central to the digital twin concept. At the same time, the approach preserves transparency and reproducibility, as it does not rely on proprietary or inaccessible clinical datasets.

It is important to emphasize that the augmented sequences are intended for illustrative and architectural validation purposes rather than clinical representation. Future work will focus on integrating real longitudinal datasets, such as continuous glucose monitoring data combined with behavioral annotations, to support clinically grounded evaluation.

Access to the OhioT1DM dataset was obtained through the official research distribution process provided by the dataset authors.

\subsection{Benchmark Models and Evaluation Metrics}
To evaluate the predictive component of the framework, four widely used regression models are considered: linear regression as an interpretable baseline, random forest regression as a nonlinear ensemble method, gradient boosting regression as a strong baseline for structured data, and a multilayer perceptron (MLP) representing neural network-based learning.

Model performance is evaluated using mean absolute error (MAE), root mean squared error (RMSE), and the coefficient of determination ($R^2$). These metrics provide complementary perspectives on prediction accuracy and error characteristics.

In addition, a binary risk formulation is derived by thresholding the target variable at its median value. For this classification task, accuracy and area under the receiver operating characteristic curve (AUC) are reported as supplementary metrics.

The selected baselines represent widely used interpretable, ensemble-based, and neural network modeling approaches in applied machine learning.~\cite{ref28,ref29,ref30,ref31}.

\subsection{Simulated Counterfactual Scenarios}
All intervention scenarios are illustrative and not derived from patient-specific physiological calibration.
Because the benchmark dataset does not include longitudinal glucose trajectories, a synthetic counterfactual scenario is constructed to illustrate the decision support functionality of the proposed framework.

Three intervention scenarios are evaluated over a 120-minute postprandial period: a baseline meal containing 60 grams of carbohydrates, a reduced-carbohydrate meal with 30 grams, and the baseline meal combined with 15 minutes of postprandial walking.

Glucose trajectories are generated using smooth parametric response functions with added stochastic variation to approximate realistic postprandial dynamics. This design enables visual comparison of intervention effects while maintaining a clear distinction between illustrative simulation and clinical validation.

The objective of this experiment is not to establish physiological accuracy, but to demonstrate how the digital twin framework supports structured comparison of alternative actions and produces interpretable, decision-relevant outputs.

The scenario design follows the principles of \textit{in silico} experimentation, in which computational models are used to evaluate plausible outcomes prior to real-world testing~\cite{ref20,ref22}.

These experimental components provide a controlled setting for validating the architectural design and computational feasibility of the proposed framework, while leaving comprehensive clinical validation as future work.
In addition to the benchmark regression experiment, a synthetic patient-level longitudinal dataset was generated to emulate CGM-style temporal patterns with carbohydrate intake, insulin exposure, and physical activity variables. This dataset was used only to test the temporal and simulation behavior of the proposed framework, not to claim clinical validity.

For additional real-world temporal analysis, CGM measurements from the OhioT1DM dataset were processed from 24 XML files across the 2018 and 2020 training and testing partitions~\cite{ref40}. A total of 166,533 CGM records were extracted, with glucose measurements sampled at approximately five-minute intervals. This analysis was used to illustrate real-world glucose dynamics and complement the simulated counterfactual scenarios.

\begin{table}[htbp]
\centering
\renewcommand{\arraystretch}{1.1}
\rowcolors{2}{rowgray}{white}
\setlength{\tabcolsep}{4pt}
\begin{tabular}{lc}
\toprule
\rowcolor{headerblue}
\textbf{Statistic} & \textbf{Value} \\
\midrule
Total CGM records & 166{,}533 \\
Number of files & 24 \\
Sampling interval & $\sim$5 minutes \\
Mean glucose (mg/dL) & 159.58 \\
Standard deviation & 60.67 \\
\bottomrule
\end{tabular}
\caption{Summary statistics of the OhioT1DM CGM dataset used in this study.}
\label{tab:ohio_summary}
\end{table}

\begin{figure}[htbp]
\centering
\includegraphics[width=0.9\linewidth]{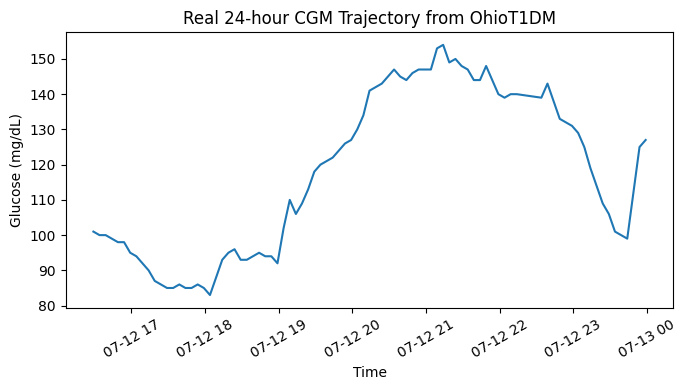}
\caption{Example 24-hour real-world continuous glucose monitoring (CGM) trajectory extracted from the OhioT1DM dataset, illustrating realistic temporal glucose dynamics.}
\label{fig:ohio_real}
\end{figure}

\section{Results}
\subsection{Benchmark Regression Performance}
Table~\ref{tab:regression} summarizes the regression performance on the benchmark diabetes dataset. Among the evaluated models, gradient boosting achieved the lowest root mean squared error (RMSE) of 53.84, closely followed by linear regression with an RMSE of 53.85. Random forest exhibited slightly lower performance, while the multilayer perceptron (MLP) showed substantially higher error and a negative $R^2$, indicating poor generalization under the selected configuration.

The small performance gap between gradient boosting and linear regression suggests that the underlying dataset exhibits limited nonlinear complexity. This is expected given the low dimensionality and static nature of the benchmark dataset. In this setting, well-regularized classical models remain competitive and provide strong, interpretable baselines for structured medical data.

\begin{table}[htbp]
\centering
\renewcommand{\arraystretch}{1.15}
\setlength{\tabcolsep}{5pt}
\rowcolors{2}{rowgray}{white}
\begin{tabular}{>{\raggedright\arraybackslash}p{3.1cm}ccc}
\toprule
\rowcolor{headerblue}
\textbf{Model} & \textbf{MAE} & \textbf{RMSE} & \textbf{$R^2$} \\
\midrule
Gradient Boosting & 44.60 & \textbf{53.84} & 0.453 \\
Linear Regression & 42.79 & 53.85 & 0.453 \\
Random Forest & 44.58 & 54.76 & 0.434 \\
MLP & 62.08 & 79.33 & -0.188 \\
\bottomrule
\end{tabular}
\caption{Regression performance on the benchmark diabetes dataset. Lower RMSE indicates better predictive accuracy.}
\label{tab:regression}
\end{table}

These results support two key observations. First, the inclusion of a transparent and reproducible benchmark provides a grounded validation of the predictive component without overstating clinical applicability. Second, the strong performance of gradient boosting and linear regression suggests that well-tuned classical models remain highly competitive for structured medical datasets and are suitable candidates for integration within digital twin frameworks, particularly for population-level risk estimation tasks.

The observed performance of gradient boosting is consistent with prior evidence that boosted tree models perform well on structured tabular data~\cite{ref28}.

\subsection{Derived Classification Benchmark}
To further evaluate predictive capability, a binary risk classification task was constructed by thresholding the target variable at its median value. Table~\ref{tab:classification} summarizes the results.

The classification results further support the use of ensemble-based models for risk stratification and anomaly-related decision tasks~\cite{ref15,ref29}.

Among the evaluated models, random forest achieved the highest area under the receiver operating characteristic curve (AUC) of 0.835 and an accuracy of 0.753. Logistic regression performed comparably, with an AUC of 0.826 and an accuracy of 0.742. Gradient boosting and the multilayer perceptron (MLP) exhibited lower performance under the selected configuration.

\begin{table}[htbp]
\centering
\renewcommand{\arraystretch}{1.15}
\setlength{\tabcolsep}{6pt}
\rowcolors{2}{rowgray}{white}
\begin{tabular}{>{\raggedright\arraybackslash}p{3.1cm}cc}
\toprule
\rowcolor{headerblue}
\textbf{Model} & \textbf{Accuracy} & \textbf{AUC} \\
\midrule
Random Forest & 0.753 & 0.835 \\
Logistic Regression & 0.742 & 0.826 \\
Gradient Boosting & 0.697 & 0.782 \\
MLP & 0.674 & 0.761 \\
\bottomrule
\end{tabular}
\caption{Classification performance using a derived binary risk label. AUC reflects model discrimination ability.}
\label{tab:classification}
\end{table}

These findings indicate that ensemble methods and generalized linear models provide robust and interpretable baselines for risk stratification tasks. This is particularly important in clinical decision support settings, where model transparency and reliability are critical.

\subsection{Counterfactual Scenario Analysis}
\begin{figure}[htbp]
\centering
\includegraphics[width=0.9\linewidth]{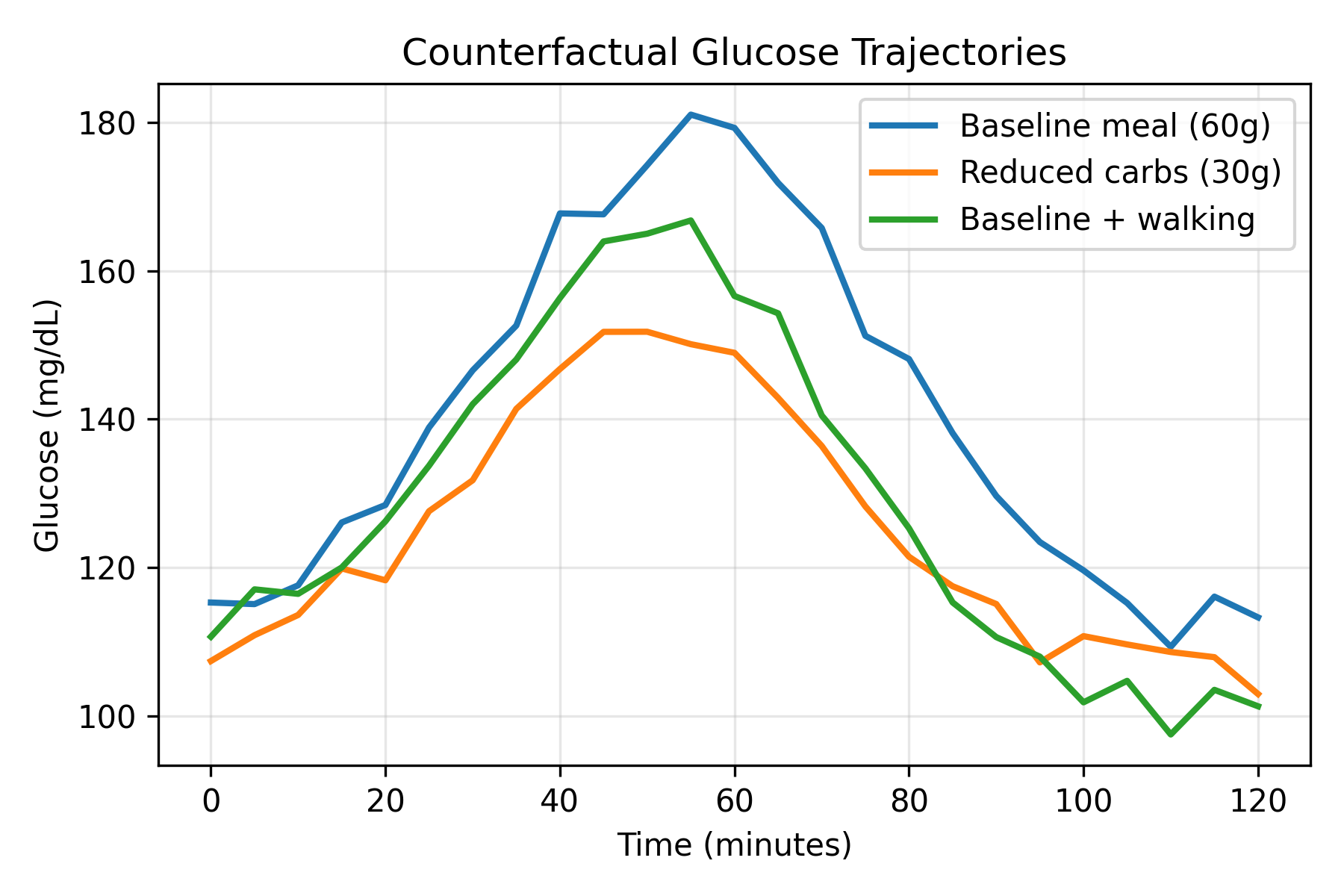}
\caption{Simulated counterfactual glucose trajectories under different intervention scenarios, illustrating the impact of reduced carbohydrate intake and postprandial physical activity on glycemic response.}
\label{fig:counterfactual_plot}
\end{figure}

The decision-support capability of the proposed framework is demonstrated through counterfactual scenario simulation. Figure~\ref{fig:counterfactual_plot} illustrates the simulated glucose trajectories under different intervention scenarios, while Table~\ref{tab:counterfactual} summarizes the corresponding quantitative outcomes.

\begin{figure}[htbp]
\centering
\includegraphics[width=0.9\linewidth]{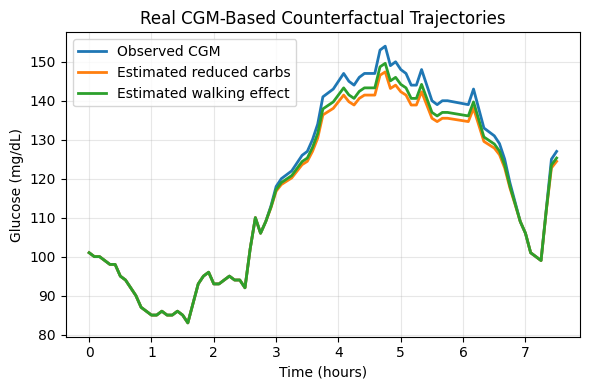}
\caption{Illustrative counterfactual trajectories derived from real CGM data. Alternative curves represent estimated effects of reduced carbohydrate intake and postprandial activity for demonstration purposes.}
\label{fig:ohio_counterfactual}
\end{figure}

Figure~\ref{fig:ohio_counterfactual} illustrates how the proposed framework may be used to explore potential intervention effects using real-world CGM data combined with simulated counterfactual trajectories.

The simulated trajectories in Fig.~\ref{fig:counterfactual_plot} provide visual evidence supporting the improvements in peak glucose and time-in-range reported in Table~\ref{tab:counterfactual}. Fig.~\ref{fig:ohio_counterfactual} further illustrates how similar intervention effects can be explored using real-world CGM data.

The reduced-carbohydrate scenario yields the largest decrease in peak glucose, lowering the maximum value from 179 mg/dL to 153 mg/dL and improving time-in-range from 58\% to 72\%. The addition of postprandial walking also reduces glucose excursion, achieving a peak of 163 mg/dL and a time-in-range of 68\%. These patterns are consistent with known physiological trends, where reduced carbohydrate intake and increased physical activity contribute to improved glycemic control.
These results should be interpreted as illustrative demonstrations of the framework rather than clinically validated outcomes.
\begin{table}[htbp]
\centering
\renewcommand{\arraystretch}{1.10}
\rowcolors{2}{rowgray}{white}
\setlength{\tabcolsep}{3pt}
\begin{tabular}{p{3cm}cc}
\toprule
\rowcolor{headerblue}
\textbf{Scenario} & \textbf{Peak Glucose (mg/dL)} & \textbf{Time-in-Range (\%)} \\
\midrule
Baseline meal (60 g) & 179 & 58 \\
Reduced carbohydrates (30 g) & 153 & 72 \\
Baseline plus walking & 163 & 68 \\
\bottomrule
\end{tabular}
\caption{Counterfactual simulation results illustrating intervention effects on peak glucose and time-in-range. Values are derived from a simulated setup.}
\label{tab:counterfactual}
\end{table}

These results demonstrate how predictive modeling can be extended to intervention analysis, which represents the central distinction between forecasting and decision support~\cite{ref18,ref19,ref33}. Rather than producing a single outcome estimate, the framework enables structured comparison of multiple intervention strategies and quantifies their relative effects.

It is important to emphasize that these values should be interpreted as qualitative demonstrations of intervention effects rather than quantitative clinical estimates. The results are derived from a simulated illustrative setup and do not represent validated clinical outcomes. Their purpose is to demonstrate how predictive modeling can be extended into actionable decision support through counterfactual reasoning.

Overall, these findings support the feasibility of the proposed architecture while highlighting the need for future evaluation on longitudinal clinical datasets.

All experimental outputs, including generated trajectories and tabular results (CSV/XML formats), are publicly available in the accompanying repository~\cite{ref41}.

\section{Discussion}
This paper presented a personalized digital twin framework for multiple diabetes types that advances artificial intelligence from predictive modeling toward decision-aware clinical support. Rather than estimating outcomes in isolation, the framework enables structured evaluation of intervention strategies through counterfactual simulation and causal reasoning. This approach provides a foundation for actionable, patient-specific decision-making in healthcare systems.

The experimental results support this transition at a conceptual level. The benchmark evaluation demonstrates that established machine learning models can provide stable predictive baselines on structured data, while the counterfactual simulation illustrates how these predictions can be extended to intervention analysis. Together, these components highlight the feasibility of integrating prediction and decision support within a unified framework.

The multi-type formulation allows the framework to adapt across clinical contexts, including Type 1, Type 2, and gestational diabetes, while preserving disease-specific constraints. This flexibility supports a range of decision scenarios, from insulin adjustment to lifestyle planning.

Importantly, the results suggest that even relatively simple predictive models can become significantly more useful when embedded within a decision-oriented architecture. The value of the system is therefore not only in predictive accuracy, but in its ability to support actionable reasoning under uncertainty.

This transition aligns with broader trends in applied machine learning toward transparent, actionable, and risk-aware decision support~\cite{ref15,ref16,ref17,ref35,ref36,ref37}. Interpretability is essential in this setting, as recommendations must be explainable and clinically meaningful.

Despite its potential, several challenges remain for real-world deployment, including data quality, generalization across populations, system interoperability, and governance concerns such as safety, privacy, and accountability~\cite{ref7,ref8,ref13,ref21,ref22,ref38}. These challenges highlight the need for further validation using longitudinal clinical data.

Clinical applicability requires validation on longitudinal patient data and remains a direction for future work.

From a broader perspective, this work reflects a conceptual shift in healthcare artificial intelligence from passive prediction toward active decision support. By enabling structured comparison of alternative interventions, the proposed framework aligns with emerging needs in precision medicine, where patient-specific decision-making is essential. This shift has implications not only for diabetes management but also for a wider class of chronic diseases that require continuous monitoring and adaptive intervention strategies.

\section{Limitations}
This study has several limitations. First, the benchmark dataset does not include continuous glucose monitoring time-series data and therefore does not support real-time glycemic forecasting. As a result, the temporal and intervention-related components are evaluated using simulated scenarios rather than real longitudinal observations.

Second, the counterfactual trajectories are generated using simplified parametric models and are intended for illustrative purposes only. While these simulations follow physiologically plausible patterns, they do not capture the full complexity of real patient responses.

Third, the causal structure is presented as a conceptual design rather than a fully estimated causal model derived from observational or interventional data. This limits the ability to draw causal conclusions in a strict statistical sense.

Fourth, no prospective clinical validation or clinician-in-the-loop evaluation is included. The framework is therefore not intended for direct clinical use in its current form.

These limitations are consistent with challenges identified in prior work on healthcare AI and digital twins, including issues related to data quality, generalization, and clinical validation~\cite{ref7,ref8,ref13,ref21,ref22}.

The objective of this study is not to establish clinical performance, but to provide a transparent and reproducible foundation for simulation-driven digital twin research in diabetes.

\section{Future Work}
Future work should evaluate the framework using longitudinal continuous glucose monitoring datasets with integrated annotations for meals, physical activity, sleep, medication, and insulin. Such data would enable direct assessment of short-term forecasting, patient-state adaptation, and intervention simulation under realistic conditions.

An important direction is the integration of mechanistic glucose–insulin models with machine learning. Hybrid approaches may improve physiological fidelity while retaining the flexibility of data-driven modeling. In addition, privacy-preserving techniques such as federated learning can support distributed model training without centralized data sharing~\cite{ref11}.

Further extensions may include constrained reinforcement learning for policy optimization, subject to strict safety constraints and clinical oversight~\cite{ref33,ref34}. Deployment studies are also needed to evaluate the impact of digital twin outputs on patient understanding, clinician trust, adherence, and clinical outcomes.

Emerging regenerative therapies, including stem cell--based approaches for pancreatic beta-cell replacement, represent a promising complementary direction for diabetes treatment. Future work may explore how digital twin models can support patient selection, risk stratification, and personalized treatment planning for such therapies. However, these approaches remain under active clinical investigation and continue to face challenges related to immune rejection, long-term graft survival, and clinical scalability~\cite{ref39}.

Finally, explainability methods should be incorporated to enhance transparency and support clinical adoption~\cite{ref35,ref36,ref37}. Gestational diabetes represents a particularly important application domain due to its stringent maternal and fetal safety requirements.

\section{Conclusion}
This paper presented a personalized digital twin framework for multiple diabetes types that emphasizes simulation-driven decision support rather than prediction alone. The framework integrates clinical, physiological, and behavioral data, maintains a patient-specific latent state, and applies counterfactual simulation to compare feasible interventions prior to real-world execution.

A reproducible proof-of-concept evaluation using an open diabetes benchmark dataset and simulated intervention scenarios demonstrates the feasibility of the proposed approach. While the results are not intended to establish clinical validation, they provide a transparent and rigorous foundation for future research on causal, interpretable, and patient-specific digital twin systems in diabetes care.

With further development using longitudinal continuous glucose monitoring data, physiological calibration, explainability methods, and prospective validation, personalized diabetes digital twins may contribute to a shift from reactive monitoring toward proactive and individualized intervention planning~\cite{ref4,ref7,ref16,ref20,ref22,ref35}. In addition, their potential integration with emerging therapeutic paradigms, including regenerative approaches such as beta-cell replacement, highlights the broader role of digital twins in supporting personalized clinical decision-making.

This shift toward decision-aware artificial intelligence enables a new class of clinical systems that move beyond passive monitoring toward proactive, personalized intervention planning.

\end{document}